\documentclass[runningheads]{llncs}

\usepackage[T1]{fontenc}
\usepackage{graphicx,verbatim}
\usepackage{amsmath,amssymb}
\usepackage{url}
\usepackage{makecell}
\usepackage{multicol,multirow}
\usepackage{xcolor}

\begin{document}
\title{Analysis of Image-and-Text Uncertainty Propagation in Multimodal Large Language Models with Cardiac MR-Based Applications}
\titlerunning{Analysis of Image-and-Text Uncertainty Propagation}
\author{
Yucheng Tang\inst{1,2} \and 
Yunguan Fu\inst{1,2} \and 
Weixi Yi\inst{1,2} \and
Yipei Wang\inst{1,2} \and
Daniel C. Alexander\inst{2,3} \and
Rhodri Davies\inst{4} \and
Yipeng Hu\inst{1,2}
}

\authorrunning{Yucheng Tang, Yunguan Fu et al.}

\institute{
Department of Medical Physics and Biomedical Engineering, University College London, UK \and
UCL Hawkes Insitute, University College London, UK \and
Department of Computer Science, University College London, UK \and
Institute of Cardiovascular Science, University College London, UK \\
\email{yucheng.tang.24@ucl.ac.uk}
}


\maketitle              
\begin{abstract}


Multimodal large language models (MLLMs) can process and integrate information from multimodality sources, such as text and images. However, interrelationship among input modalities, uncertainties due to individual uni-modal data and potential clinical applications following such an uncertainty decomposition are yet fully understood in the context of large-scale MLLMs.
In this work, we propose a multimodal uncertainty propagation model (MUPM) based on uncertainty propagation, to characterise the relationship among the uncertainties arising from image-only, text-only, and joint image-text variations in MLLM inputs. Using real clinical data consisting of cardiac MR scans and digital health records, we describe that MUPMs can be optimised robustly with a few samples. We then show that the fitted MUPMs are generalisable across different input data distributions and, perhaps surprisingly, across different downstream tasks. Such a transferability may be explained by the shared pretraining, comparatively light MLLM fine-tuning, along with the low-dimensional nature of the MUPMs. More importantly, this learned transferability, quantifying the relationship between these uncertainties, led to direct clinical applications in which uncertainties may be estimated and thus analysed robustly for varying data or even a novel set of cardiac disease prediction tasks.
In addition, we show experimentally the efficiency in multimodal data required for estimating the overall uncertainty and its ability to identify redundant factors, both of which are considered practical yet clinically useful applications with the proposed MUPMs.
Codes are available at \url{https://github.com/yucheng722/MUPM}.

\keywords{Uncertainty \and Multimodal LLM \and Cardiac MRI}

\end{abstract}
\section{Introduction}

By leveraging multimodality inputs such as images and texts, multimodal large language models (MLLMs) such as BLIP-2 \cite{li2023blip}, LLaVA-Med \cite{li2024llava}, and M3D \cite{bai2024m3d} have achieved state-of-the-art performance in tasks like clinical decision-making \cite{li2023meddm}, medical image processing \cite{chen2024trimedlm}, and diagnostic text generation \cite{mo2024large}. Analysis of uncertainties from individual modality data and their interactions in MLLMs remain underexplored.
In medicine, where such uncertainty analysis is particularly useful as it directly impacts the credibility of clinical decisions and potentially offers quantitative insights in these decisions, interesting questions arise: How does variability in a single modality affect overall uncertainty? How can we utilize these uncertainties for clinical applications?

Previous work leveraged MLLM uncertainty to enhance performance in specific tasks, such as composite image retrieval \cite{ge2025llm}, out-of-distribution detection \cite{dai2023exploring}, and hallucination identification \cite{zhang2024vl}. Other studies investigated calibration before and after fine-tuning \cite{chen2024unveiling}, relationship between accuracy and uncertainty \cite{kostumov2024uncertainty}, and uncertainty-based evaluation for MLLMs \cite{dang2024exploring}. However, to our best knowledge, little research has studied relationship between uncertainties across different MLLM modalities and the clinical applications of these relationships.

In this work, we propose the multimodal uncertainty propagation model (MUPM) to characterize the relationship among uncertainties from image-only, text-only, and joint image-text inputs, as well as the correlation between image and text inputs, in MLLMs. We refer hereinafter to these uncertainties as image-only, text-only, and overall uncertainties.
We apply the MUPM to various cardiac disease prediction applications, using cardiac magnetic resonance (MR) imaging scans combined with patient digital health records. 
We adopted the existing medical-domain-specific M3D model \cite{bai2024m3d}, which has been proven credible on large biomedical datasets and supports 3D medical image inputs.
Subsequently, we investigate a set of MUPM applications including its robustness across different input data distributions and downstream tasks, efficient estimation of overall uncertainty, and identifying redundant factors.

Pioneering analysis of multi-modality uncertainty propagation, our contributions are as follows:
First, we derive a mathematical expression to quantify the relationship among image-only uncertainty, text-only uncertainty, the correlation between image and text inputs, and the overall uncertainty in MLLMs.
Secondly, we fit MUPM instead of computing derivatives and show experimentally improved stability and generalizability. Furthermore, we identify four novel applications of MUPM for cardiac disease prediction, with the experiments presented using large data sets. We release our code at \url{https://github.com/yucheng722/MUPM}.

\section{Method}
\subsection{First-order Variances and Covariance Propagation}
Uncertainty propagation theory provides a framework for quantifying how uncertainties (or random errors) in input variables affect a model's uncertainty \cite{lee2009comparative}. 
Building on this foundation, we first examine how to quantify the overall uncertainty as a function of image-only and text-only uncertainties.

Suppose that the MLLM is a high-dimensional, nonlinear model $F$ with paired input variables: image $x$ and text $y$. A first-order Taylor expansion approximates the model in Eq.~(\ref{Taylor}).
\begin{equation}
F(x,y) \approx F(x_0,y_0) + \frac{\partial F}{\partial x} x + \frac{\partial F}{\partial y} y
    \label{Taylor}
\end{equation}
where $\partial F/\partial x$ and $\partial F/\partial y$ are the partial derivatives of the model with respect to image and text inputs, respectively. By computing the variance of both sides of Eq.~(\ref{Taylor}), the uncertainty propagation formula is derived as in Eq.~(\ref{Var}).
\begin{equation}
\sigma_{F_{I,T}}^2 \approx \Big(\frac{\partial F}{\partial x}\Big)^2 \sigma_I^2 + \Big(\frac{\partial F}{\partial y}\Big)^2 \sigma_T^2 + 2 \Big(\frac{\partial F}{\partial x}\Big)\Big(\frac{\partial F}{\partial y}\Big)\rho\sigma_{I}\sigma_T
    \label{Var}
\end{equation}
where $\sigma_I^2$ and $\sigma_T^2$ represent the variances of the image and text inputs, respectively. 
$\rho \sigma_I \sigma_T$ represents the covariance between image and text inputs, where $\rho$ is the Pearson correlation coefficient~\cite{cohen2009pearson}. In subsequent sections, we refer to $\sigma_I \sigma_T$ as the covariance term.
$\sigma_{F_{I,T}}^2$ denotes the variance of the model's output.

By substituting $\sigma_{F_I}^2 \approx a_I^2 \sigma_I^2$ and $\sigma_{F_T}^2 \approx b_T^2 \sigma_T^2$, where $a_I^2 = \left( \frac{\partial F}{\partial x} \right)^2$ and $b_T^2 = \left( \frac{\partial F}{\partial y} \right)^2$, into Eq.(\ref{Var}), the overall uncertainty can be quantified as a function of image-only and text-only uncertainties, as shown in Eq.~(\ref{Image-Text})
\begin{equation}
    \sigma_{F_{I,T}}^2 \approx \frac{a^2}{a_I^2}\sigma_{F_{I}}^2 + \frac{b^2}{b_T^2}\sigma_{F_{T}}^2 + 2 \frac{ab}{a_Ib_T} \rho \sigma_{F_I}\sigma_{F_T}
    \label{Image-Text}
\end{equation}
where $a$ and $b$ denote the partial derivatives of the model with respect to $x$ and $y$. 
Explicit computing these partial derivatives is challenging due to the complex, nonlinear architecture of MLLMs and is numerically unstable, as shown in experiments of instability of derivative computation in Section 3, whereas fitting linear models provides a more reliable and computationally feasible alternative for approximating these derivatives.

\subsection{Uncertainty Estimation and MUPM Fitting}
\begin{figure}[t]
    \centering
    \includegraphics[width=\textwidth,height=4cm]{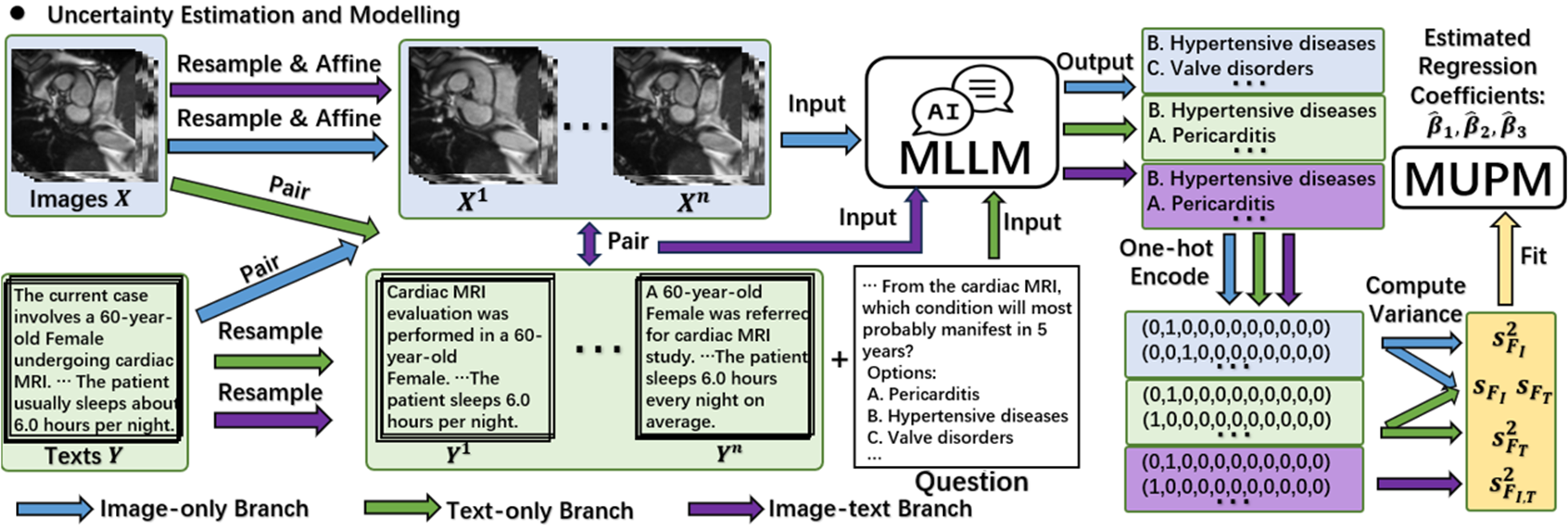}
    \caption{Uncertainty estimation and MUPM fitting process. 
 We compute the sample variances of one-hot encoded outputs of MLLM when applying image-only, text-only, and image-text data resampling and augmentation. The covariance term is computed from the outputs of the image-only and text-only branches. 
    We then fit the linear MUPM in Eq.~(\ref{MUPM}) using the sample variances and the covariance term. The regression coefficients are estimated during this process.}
    \label{Fig1}
\end{figure}

To estimate uncertainties in Eq.~(\ref{Image-Text}), we use resampling and augmentation strategies \cite{wang2019aleatoric} to approximate uncertainties without retraining the MLLM. 
MUPM is not an uncertainty quantification method, but a model-agnostic framework and is compatible with various uncertainty estimation approaches, such as infer-dropout \cite{mi2022training}, greedy negative log-likelihood (G-NLL) \cite{aichberger2024rethinking}, named entity replacement \cite{goodarzi2023robustness}, and assembly \cite{he2020towards}.

Let $X\text{-}Y$ be a paired image-text input set, where $X = \{x_i\}_{i=1}^N$ represents $N$ image inputs and $Y = \{y_i\}_{i=1}^N$ represents the corresponding text inputs. Each pair $(x_i, y_i)$ describes the same semantic entity. As shown in Fig.~\ref{Fig1}, to estimate the image-only uncertainty $\sigma_{F_I}^2$, we generate $n$ augmented versions for each cardiac MR image in the original set $X$. These augmented images, denoted by $\{X^{j}\}_{j=1}^n$, are produced by applying affine transformations (e.g., rotation, zoom, shear, shift) and adding Gaussian noise to resampled scans from the same patient case. We pair these images with fixed text inputs $Y$ and compute the sample variance $s_{F_I}^2$ from the one-hot encoded MLLM outputs. One-hot encoding of MLLM outputs is application-specific; further details are provided in Section 3.3. Similarly, $\sigma_{F_T}^2$ is estimated by augmenting resampled digital health records $\{Y^{j}\}_{j=1}^n$ using synonymous replacements while fixing image inputs $X$, then calculating sample variance $s_{F_T}^2$ of the one-hot encoded MLLM outputs. 
For overall uncertainty $\sigma_{F_{I,T}}^2$, we simultaneously resample and augment both modalities, pair augmented samples, and compute MLLM outputs' variance $s_{F_{I,T}}^2$.

Based on Eq.(\ref{Image-Text}), we formulate MUPM as a linear regression model in Eq.(\ref{MUPM}), with independent variables ($s_{F_I}^2$, $s_{F_T}^2$, $s_{F_I}s_{F_T}$), and dependent variable ($s_{F_{I,T}}^2$).
\begin{equation}
s_{F_{I,T}}^2 = \beta_1 s_{F_I}^2 + \beta_2 s_{F_T}^2 + \beta_3 s_{F_I}s_{F_T} + \epsilon
    \label{MUPM}
\end{equation}
where $\beta_1$, $\beta_2$, and $\beta_3$ are regression coefficients, $\epsilon$ represents residual errors. The least squares~\cite{bjorck1990least} is used to estimate optimal parameters $\hat{\beta}_1$, $\hat{\beta}_2$, and $\hat{\beta}_3$.

\subsection{MUPM Applications}
We propose four clinical use cases of MUPM for cardiac disease prediction. 

First, MUPM is investigated across various input data distributions for cardiac disease prediction. For instance, if we observe a higher regression coefficient of text-only (or image-only) uncertainty in MUPM, meaning that it contributes more to the overall uncertainty, we would like to improve the text (or image) quality by e.g. adding more detailed or further clinical information (or by MR reacquisition). In these examples, if the quantified relationship between these uncertainties remains relatively stable, there is no need to refit MUPM. 
It is nonetheless important to note that quantifying the health-economical values of per unit of changes in texts or images remains highly application-specific. Specific clinical cases are provided and discussed in Section 3.


Second, MUPM is tested with various cardiac disease prediction tasks, including forecasting the occurrence of diseases in 1, 3, or 5 years, using closed-ended visual question answering (VQA) tasks in which MLLM classifies diseases from predefined options. 
These tasks are designed to share the same definition of uncertainty, as the same token-based format encoded in one-hot is used, even though they produce different semantic predictions.

Furthermore, MUPM is expected to enable efficient estimation of overall uncertainty from downstream data with fewer resample sizes ($n$ in Section 2.2 set to a lower value) than estimation using test-time augmentation from scratch.

Finally, MUPM identifies redundant factors by analyzing the covariance term. If a modality has a very low regression coefficient, it has a low impact on predictive performance. Completely redundant factors are more likely to have very low regression coefficients and simultaneously exhibit elements with large absolute values in their covariance terms with other modalities.

\section{Experiments and Results}
\textbf{Implementation Details}
We implemented MLLM uncertainty estimation using PyTorch \cite{paszke2019pytorch} and Transformers \cite{wolf2020transformers}, and fitted MUPM using Scikit-Learn \cite{pedregosa2011scikit}. We adopted the M3D \cite{bai2024m3d} model 
as the MLLM model as it supports 3D image inputs.
The implementation was carried out on a 16GB NVIDIA GeForce RTX 4070 GPU, and the uncertainty estimation and MUPM fitting process took approximately 8 hours in total.

\textbf{Datasets}
We utilized the UK BioBank Dataset Heart MRI (Category 1015
, Version Sep 2024) as image database, which includes data from 78,684 participants, resulting in 71,702 image pairs. These images include four cardiac MRI modalities: lax2c, lax3c, lax4c (2D long-axis views), and sax (3D short-axis view), each with 50 temporal frames.
For text database, we collected information from the UK Biobank Text Metadata (Data-Field 41270
, Data-Field 41280
, Version Apr 2024), covering 446,814 participants and totaling 7,015,683 entries. The records include 17 patient information items such as age, sex, weight, height, BMI, symptoms, pulse rate, diastolic blood pressure, systolic blood pressure, cardiac related history, cardiac history, family history, metabolic equivalent of task, sleep duration, smoking, drinking, heart function and diagnoses of 11 cardiac disease categories: hypertensive disease, ischaemic heart disease, arrhythmia, conduction disorders, complications, valve disorders, heart failure, pulmonary heart disease, rheumatic heart disease, pericarditis, and cardiomyopathy.

We constructed image-text pairs by retrieving cardiac disease diagnostic records from the text database, aligned with each patient’s cardiac MRI timestamp at 1, 3, and 5 year intervals. This yielded $611$, $1223$, and $1223$ valid cardiac image-text pairs (for 1, 3, and 5 year predictions, respectively) for a closed-ended VQA task predicting cardiac diseases.
The original images were resized to $256\times256\times32$ to meet the M3D model’s input requirements and normalized to $[0, 1]$. In all simulations, we set the resample size $n = 20$.
We calculated the benchmark of overall uncertainty using all cardiac image-text pairs, which includes cardiac MR images and health record text pairs used for cardiac disease prediction over 1, 3, and 5 years. Resampling and augmentation were applied 100 times to quantify overall uncertainties for all cardiac image-text pairs, and their average was used as the benchmark for comparison.

To avoid \textbf{instability of derivative computation}, also described in Section 2.1 and Section 2.2, we use the linear regression model in Eq.(\ref{MUPM}), where the fitted regression coefficients represent approximations of the complex partial derivative-related terms in Eq.(\ref{Image-Text}).
 To validate this, $60$ cardiac image-text pairs were sampled from our dataset, to estimate the partial derivatives using the chain rule, for all three types of uncertainties. For each sampled cardiac image-text pair, we computed the average derivatives of image pixels (for cardiac MR images) and text tokens (for health record texts). 
The results are as follows: $\overline{a}_I=4.58 \pm 5.69$, $\overline{b}_T = 5.65 \pm 5.79$, $\overline{a} = 3.16 \pm 5.33$, and $\overline{b} = 5.07 \pm 5.91$, where $\overline{a}_I$, $\overline{b}_T$, $\overline{a}$, and $\overline{b}$ represent the computed average derivatives with standard deviations (StDs).
The results indicate that directly computing these partial derivatives yields large standard deviations among samples, highlighting their instability. Additionally, gradient vanishing was observed in this experiment, making uncertainty propagation modelling highly impractical.



\begin{table}[t]
\centering
\caption{Results of MUPM regression when input data distributions change due to improved data quality in cardiac disease prediction tasks. 
$\hat{\beta}_1$, $\hat{\beta}_2$, and $\hat{\beta}_3$ are the averaged fitted regression coefficients from five experiments, corresponding to the L2-norms of the mean cardiac MR image-only uncertainty ($\|s^2_{F_I}\|_2$), health record text-only uncertainty ($\|s^2_{F_T}\|_2$), and the covariance term ($\|s_{F_I}s_{F_T}\|_2$).  $R^2$ represents goodness of fit between the overall uncertainties computed by the fitted MUPM and overall uncertainty benchmark, with reported ECE values and the ANOVA p-values for the regression coefficients across the three tasks.
The abbreviations denote experimental settings: w.o. (without data quality improvement); Image
$\uparrow$ (cardiac MR image quality improvement only); 
Text $\uparrow$  (health record text quality improvement only); Both $\uparrow$  (simultaneous cardiac MR image and health record text quality improvement).
}
\setlength{\tabcolsep}{13pt} 
\resizebox{\textwidth}{!}{
\begin{tabular}{c|ccc|cc}
\hline
& $\hat{\beta}_1\;(\|s^2_{F_I}\|_2)$ & $\hat{\beta}_2\;(\|s^2_{F_T}\|_2)$ & $\hat{\beta}_3\;(\|s_{F_I}s_{F_T}\|_2)$ & $R^2$ & ECE  \\
\hline\hline
w.o. & $0.24\;(0.18)$ & $0.90\;(0.40)$ & $-0.20\;(0.17)$ & $0.76$ & $0.09$\\
Image $\uparrow$& $0.24\;(0.15)$ & $0.88\;(0.42)$ & $-0.21\;(0.16)$ & $0.78$ & $0.09$\\
Text $\uparrow$& $0.21\;(0.18)$ & $0.93\;(0.35)$ & $-0.22\;(0.13)$ & $0.80$ & $0.12$ \\
Both $\uparrow$ & $0.22\;(0.11)$ & $0.91 \;(0.28)$ & $-0.22 \;(0.09)$ & $0.84$ & $0.06$\\
\hline
ANOVA & $0.41$ & $0.33$ & $0.89$
\\
\hline
\end{tabular}}
\label{biao3}
\end{table}

\begin{table}[t]
\centering
\caption{Results of MUPM regression across different downstream tasks for predicting cardiac disease over 1, 3, 5 years, respectively, structured the same as Table~\ref{biao3}.}
\setlength{\tabcolsep}{13pt} 
\resizebox{\textwidth}{!}{
\begin{tabular}{c|ccc|cc}
\hline
& $\hat{\beta}_1\;(\|s^2_{F_I}\|_2)$ & $\hat{\beta}_2\;(\|s^2_{F_T}\|_2)$ & $\hat{\beta}_3\;(\|s_{F_I}s_{F_T}\|_2)$ & $R^2$ & ECE\\
\hline\hline
1 year& $0.20 \;(0.11)$ & $0.91 \;(0.30)$ & $-0.19 \;(0.09)$ & $0.79$ & $0.07$ \\
3 years& $0.21\;(0.12)$ & $0.89 \;(0.32)$ & $-0.20 \;(0.10)$ & $0.82$  & $0.07$ \\
5 years& $0.23\;(0.13)$ & $0.91 \;(0.33)$ & $-0.22 \;(0.10)$ & $0.78$ & $0.10$\\
\hline
ANOVA & $0.70$ & $0.95$ & $0.98$ & \\
\hline
\end{tabular}}
\label{biao1}
\end{table}

\begin{figure}[t]
    \begin{minipage}[t]{0.55\linewidth}
        \centering
        \includegraphics[width=\textwidth,height = 4.5cm]{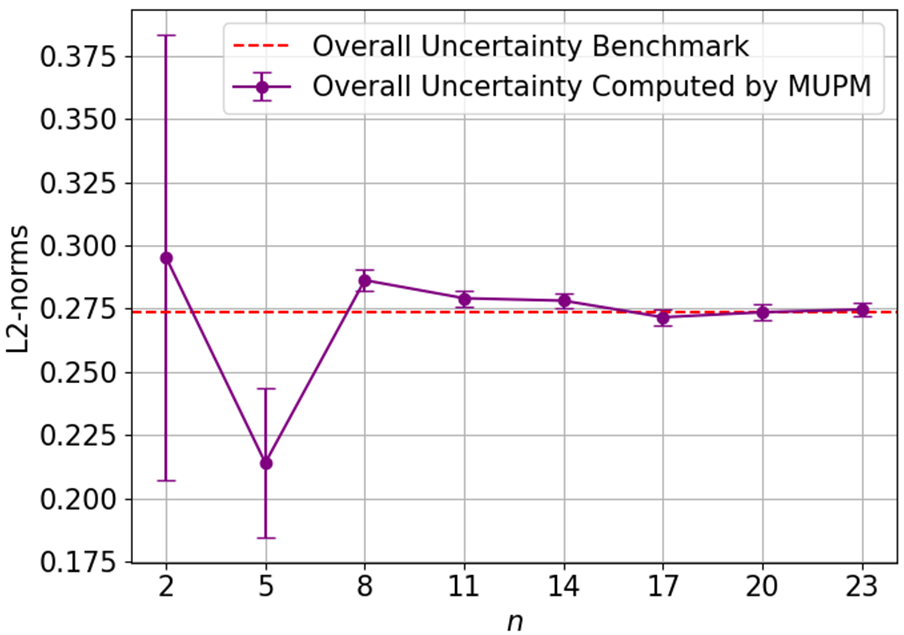}
        \centerline{(a)}
        \label{Exp3}
    \end{minipage}%
    \begin{minipage}[t]{0.40\linewidth}
        \centering
        \includegraphics[width=\textwidth,height = 4.5cm]{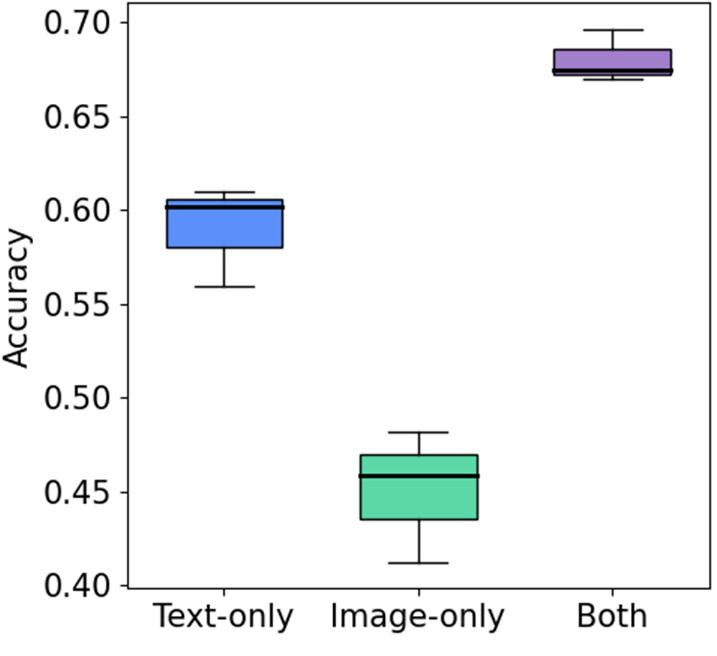}
        \centerline{(b)}
        \label{Exp4}
    \end{minipage}
    \caption{(a) Experimental results on the efficient uncertainty estimation of downstream data: the purple line represents the average L2-norms of the overall uncertainties (with StDs) computed by MUPM, when $n = 2,5,8, 11,14,17,20,23$. The red line represents the benchmark overall uncertainty. (b) Box plot of the average accuracy (with StDs) when removing cardiac MR image inputs (text-only), removing health record text inputs (image-only), and keeping both inputs (both).}
\end{figure}

\textbf{Robustness of MUPM across various input data distributions} was quantified in the cardiac disease prediction tasks.
We used cardiac MR images in short-axis (sax) view as image inputs. 
To generate augmented cardiac MR sax image inputs for each patient case, we first resampled from a total of 50 time frames along the temporal axis. Using the proposed sampling procedure, we expected to cover all time frames across a cardiac cycle, thus representing a time-frame independent analysis. Each resampled image was then augmented using random affine transformations with maximum rotations of $(15, 15, 15)$, zooms of $(0.2, 0.2, 0.2)$, shear transformations of $(5, 5, 5, 5, 5, 5)$, and shifts of $(0.15, 0.15, 0.15)$, and an added Gaussian noise (StD=$10$). 
To generate augmented text inputs, for each input, we resampled 5 to 6 items from the patient information and combined them with the predefined options corresponding to 11 cardiac disease categories. Additionally, we used synonymous replacements by GPT \cite{achiam2023gpt} with the temperature parameter set to 0.8. The one-hot encoded prediction was one of the 11 cardiac disease categories.

To simulate real-world scenarios of cardiac MR images with different imaging angles and signal-to-noise ratios (SNR), we changed the data distributions and improved image quality (relative to those in the experiments in Table~\ref{biao1}) by resampling the cardiac MR images in long-axis (lax2c, lax3c, lax4c) views with higher SNRs and by removing affine transformations and noise.
To simulate real-world scenarios of health reports with varying writing styles, formats, and templates, for health record text inputs, we changed the data distribution by adding more patient information items (increasing from 5–6 items to 10–12 items) and employing a data augmentation strategy involving synonymous replacements generated by GPT \cite{achiam2023gpt} with a lower temperature parameter of 0.4.

For each data distribution, the cardiac image-text pairs were split into five non-overlapping folds, and MUPM was fitted using each fold for 5 times. Although quantifying uncertainty is not the main focus of this work, its quality is relevant in real-world applications.
We computed the expected calibration errors (ECEs)~\cite{nixon2019measuring} to assess the quality of the estimated uncertainty. Samples were grouped based on the $10$ percentiles of their uncertainty L2-norms. Within each group, we computed the mean absolute error (MAE) between prediction accuracy and confidence, with confidence quantified by mapping the uncertainty L2-norms using a logarithmic and sigmoid transformation. 

The experimental results based on analysis of variance (ANOVA) in Table~\ref{biao3} show that there is no evidence to reject the null hypothesis that there is no significant difference between these regression coefficients of MUPM, across different input data distributions. This demonstrates that MUPM is insensitive to variable data distribution. Notably, improving the quality of health record text alone resulted in a more significant reduction in overall uncertainty as shown in Table~\ref{biao3}. This is because the regression coefficient for text-only uncertainty is relatively large, perhaps suggesting it contributes more to overall uncertainty. 

\textbf{Robustness of MUPM in different cardiac disease predication tasks} was also tested, for predictions over 1 year, 3 years, and 5 years, respectively, with the same image-text pairs. 
Table~\ref{biao1} compares MUPM's regression coefficients across tasks, with statistical significance assessed by ANOVA\cite{st1989analysis} (p=0.05), with all pair-wise p-values greater than $0.05$. This may seem counter-intuitive, albeit also demonstrated by the absolute values, the fitted variance and covariance are relatively invariant to varying tasks.
Additionally, the results in Table~\ref{biao1} show that the average L2 norm of uncertainty increases with prediction time: 1 year < 3 years < 5 years, indicating a decrease in the reliability of cardiac disease predictions as time progresses, which is consistent with intuition in this case.

\textbf{Efficient use of downstream data in overall uncertainty estimation} was assessed by testing whether stable overall uncertainty estimates can be obtained with limited resample sizes ($n$ in Section 2.2), on unseen downstream data. By varying $n$, we analyzed the trend of overall uncertainty estimates. 
For each $n$, we sampled $60$ unseen image-text pairs to estimate the image-only and text-only uncertainties by respectively resampling and augmenting input pairs $n$ times. The average L2-norms of $60$ overall uncertainty estimates (with StDs) are shown in Fig.~2.(a) when $n=2,5,8, 11,14,17,20,23$. 
Results indicate that as $n$ increases, the overall uncertainty calculated by MUPM becomes closer to the benchmark. Beyond $n=20$, the MUPM calculated overall uncertainty stabilizes with low StDs, allowing reliable computation using MUPM. 

\textbf{Identifying Redundant Factors}
As shown in Table~\ref{biao1} and Table~\ref{biao3}, although the cardiac MR image-only regression coefficients ($\hat{\beta}_1$) are relatively small, the low covariance term between the cardiac MR image-only and health record text-only uncertainties (i.e., $ \hat{\beta}_3 \times s_{F_I}s_{F_T}$) suggests that cardiac MR image information is not redundant. To further validate this, we removed the cardiac MR image and health record text inputs separately, which led to a significant decrease in prediction accuracy (see Fig.~2.(b)), confirming that both modalities are essential in cardiac disease prediction tasks.

\section{Conclusion}
We present MUPM, a linear model that describes uncertainty propagation in MLLM. We fit MUPM by estimating the cardiac MR image-only, health record text-only, and overall uncertainties, along with the covariance term, using resampling and augmentation strategies, to avoid the instability observed in derivative computations.
Four clinical applications of MUPM are proposed for cardiac disease prediction. Experimental results show MUPM's robustness across different input data distributions and downstream tasks, in addition to efficient estimation and the ability to identify redundancy. Future work will extend MUPM to cardiac MRI–specific MLLMs such as CineMA\cite{fu2025cinema}, and apply it to per-disease analyses to identify clinically relevant modality contributions.

\subsubsection{Disclosure of Interests.} The authors have no competing interests to declare that are relevant to the content of this article.


%
%
%
\bibliographystyle{splncs04}
\bibliography{Paper-4581}

\end{document}